# Determining the Consistency factor of Autopilot using Rough Set Theory

Anugrah Kumar
School of Computing Science and Engineering
VIT University
Vellore, India
anugrah18@gmail.com

Dr. R Manjula
School of Computing Science and Engineering
VIT University
Vellore, India
rmanjula@vit.ac.in

*Abstract*— **Autopilot is a system designed to guide a vehicle without aid. Due to increase in flight hours and complexity of modern day flight it has become imperative to equip the aircrafts with autopilot. Thus reliability and consistency of an Autopilot system becomes a crucial role in a flight. But the increased complexity and demand for better accuracy has made the process of evaluating the autopilot for consistency a difficult process .A vast amount of imprecise data has been involved. Rough sets can be a potent tool for such kind of Applications containing vague data. This paper proposes an approach towards Consistency factor determination using Rough Set Theory. The seventeen basic factors, that are crucial in determining the consistency of an Autopilot system, are grouped into five Payloads based on their functionality. Consistency Factor is evaluated through these payloads, using Rough Set Theory. Consistency Factor determines the consistency and reliability of an autopilot system and the conditions under which manual override becomes imperative. Using Rough set Theory the most and the least influential factors towards Autopilot system are also determined.**

## I. Introduction

Since the inception of aviation industry, the flight security measures played an important role [1]. Traditional flight security measures involved a continuous attention of the pilot during the whole flight. Increase in flight hours led to the development of autopilot systems in flight to reduce the workload and fatigue of pilots during flights [2]. Autopilot system is designed to perform most of the tasks which would assist pilot during their flights. It directly contributes towards the safety and efficiency of a flight. The advancement in technology has made the aircraft control system more complex. Thus autopilots have become more sophisticated compared to the ones used in early 20$^{th}$ century. The datasets used to process the reliability of autopilot has increased tremendously making the data more vague and imprecise. Thus determining the consistency and reliability of an autopilot system requires a more intricate and intelligent method [3].

In this paper we analyze the vagueness of factors involved in determining the consistency of an autopilot system. This paper proposes a Rough Set theory based method to determine a consistency factor of an Autopilot. The Airbus 320 model is used for analyzing the factors responsible for the appraisal of autopilot system's consistency in a flight. The Rough Set theory is selected for this purpose since it handles vague data problems with efficiency and accuracy.

## II. Basic components of autopilot system

The Basic Components involved in Autopilot are listed below.

### A. Mode Selector

This allows autopilot to incorporate the autopilot with other modules of an aircraft. Through this the pilot can also program the anticipated flight profile.

### B. Computer

This represents the core component of an autopilot system. The computer receives and processes the information and data input from incorporated avionics components. It also sends processed signals to system's actuators.

### C. Actuators

It receives the processed computer signal and moves the control surface to achieve the intended function.

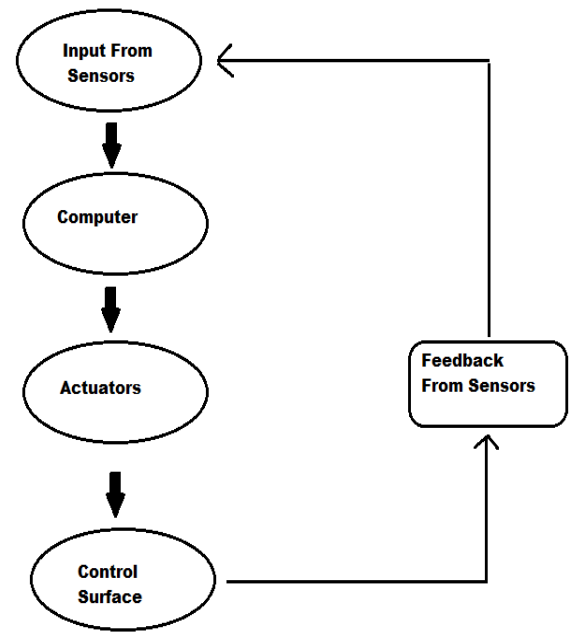

FIG. 1. Basic Autopilot Control system.

## III. ROUGH SET THEORY

Rough set is a new mathematical approach introduced by Zdzislaw Pawlak towards imprecise or vague data [3]. As safety is of prime importance during flight and dataset for evaluating autopilot system's consistency has become more vague and imprecise. Thus there is a need for a more better and intelligent method to evaluate the consistency of autopilot system during a flight. Rough Set deals with vague data efficiently. Thus this paper proposes a rough set theory approach to determine the consistency of an autopilot system.

Let 'U' be a Universal set and 'R' be an indiscernibility relation R such that $R \subseteq U \times U$ which represents lack of information about the elements of U. Let 'X' be a subset of U.

Using Basic Rough Set theory concepts

*R-Lower approximation of X*
$R_*(x) = \bigcup_{x \in U} \{R(x): R(x) \subseteq X\}$

*R- Upper approximation of X*
$R^*(x) = \bigcup_{x \in U} \{R(x): R(x) \cap X \neq \phi\}$

*R-Boundary region of X*
$RN_R(X) = R^*(x) - R_*(x)$

If $RN_R(X) = \phi$ *(Set X is Crisp Set)*
If $RN_R(X) \neq \phi$ *(Set X is Rough Set)*

The initial dataset contains impreciseness and redundancy. Thus many of its redundancy can be removed by calculating Reduct of the sets. [3]

Also
Core (X) = $\cap$ Red (X) where Core of X is set of all indispensible attributes of X. [3]

Information is represented in form of two attributes Condition and Decision attributes. All the inputs or parameters comprises of Conditions whereas Output comprises of Decision. Each row determines a decision rule. A set of Decision rules is known as Decision Algorithm. After Determining the Reduced set the respective Decision Algorithm is calculated [3].

Positive Region of U/B with respect to A is the set of all elements of U that can be uniquely classified to blocks of partition U/D by means of A [4].
$POS_A (B) = \bigcup_{X \in U/I(D)} A_*(X)$

## IV. ROUGH SET THEORY approach for determining the consistency of autopilot system

Using Consistency Factor the reliability and consistency of an Autopilot system can be determined. Basic approach involved here is to apply Rough Sets to determine the Consistency factor. There are many factors which are crucial in deciding the autopilot system's consistency. Here seventeen basic factors will be considered that play significant role in determining Consistency Factor. The seventeen factors are further stratified into five payloads based on the functionality of its components. Payload contains the dataset directly fed from various component sensors.

**Payload I:-**
1. Roll inconsistency.
2. Pitch inconsistency.
3. Yaw inconsistency.

**Payload II:-**
1. Altitude inconsistency.
2. Longitude inconsistency.
3. Latitude inconsistency.

**Payload III:-**
1. Distance Measuring equipment fault.
2. VHF Omnidirectional range fault.
3. Inertial reference systems fault.

**Payload IV:-**
1. Gyroscope instrument Failure.
2. Accelerometers instrument Failure.
3. Altimeters instrument Failure.
4. Compass instrument Failure.

**Payload V:-**
1. Flight Route Change.
2. Flaps Failure.
3. Fuel consumption inconsistency
4. Inflight Icing.

These five payloads will be analyzed independently for any irregularities and then the results of these five payloads will be combined to evaluate the consistency factor using Rough Set.

*A. Analyzing Payload I*
The Table I consists of three parameters and it represents Payload I. These parameters can have two input values yes or no. Depending upon these values on the parameters, the output known as the Payload I Consistency is generated.

TABLE I

| Roll inconsistency | Pitch inconsistency | Yaw inconsistency | Payload I Consistency |
|---|---|---|---|
| Yes | Yes | Yes | High |
| Yes | Yes | No | Moderate |
| Yes | No | Yes | Moderate |
| Yes | No | No | Low |
| No | Yes | Yes | Moderate |
| No | Yes | No | Moderate |
| No | No | Yes | Low |
| No | No | No | Extremely Low |

*B. Analyzing Payload II*
The Table II consists of three parameters and it represents Payload II. These parameters can have two input values yes or no. Depending upon these values on the parameters, the output known as the Payload II Consistency is generated.

TABLE II

| Altitude inconsistency | Longitude inconsistency | Latitude inconsistency | Payload II Consistency |
|---|---|---|---|
| Yes | Yes | Yes | High |
| Yes | Yes | No | Moderate |
| Yes | No | Yes | Moderate |
| Yes | No | No | Low |
| No | Yes | Yes | Moderate |
| No | Yes | No | Moderate |
| No | No | Yes | Low |
| No | No | No | Extremely Low |

*C. Analyzing Payload III*
The Table III consists of three parameters and it represents Payload III. These parameters can have two input values yes or no. Depending upon these values on the parameters, the output known as the Payload III Consistency is generated.

TABLE III

| Distance Measuring equipment fault. | VHF Omnidirectional range fault. | Inertial reference systems fault. | Payload III Consistency |
|---|---|---|---|
| Yes | Yes | Yes | High |
| Yes | Yes | No | Moderate |
| Yes | No | Yes | Moderate |
| Yes | No | No | Low |
| No | Yes | Yes | Moderate |
| No | Yes | No | Moderate |
| No | No | Yes | Low |
| No | No | No | Extremely Low |

*D. Analyzing Payload IV*
The Table IV consists of four parameters and it represents Payload IV. These parameters can have two input values yes or no. Depending upon these values on the parameters, the output known as the Payload IV Consistency is generated.

TABLE IV

| Gyroscope instrument Failure | Accelerometers instrument Failure | Altimeters instrument Failure | Compass instrument Failure | Payload IV Consistency |
|---|---|---|---|---|
| Yes | Yes | Yes | Yes | High |
| Yes | Yes | Yes | No | Moderate |
| Yes | Yes | No | Yes | Moderate |
| Yes | Yes | No | No | Low |
| Yes | No | Yes | Yes | Moderate |
| Yes | No | Yes | No | Low |
| Yes | No | No | Yes | Low |
| Yes | No | No | No | Extremely low |
| No | Yes | Yes | Yes | Moderate |
| No | Yes | Yes | No | Low |
| No | Yes | No | Yes | Low |
| No | Yes | No | No | Extremely low |
| No | No | Yes | Yes | Low |
| No | No | Yes | No | Extremely Low |
| No | No | No | Yes | Extremely low |
| No | No | No | No | Extremely low |

*E. Analyzing Payload V*
The Table V consists of four parameters and it represents Payload V. These parameters can have two input values yes or no. Depending upon these values on the parameters, the output known as the Payload V Consistency is generated.

TABLE V

| Flight Route Change | Flaps Failure | Fuel consumption inconsistency | Inflight Icing | Payload V Consistency |
|---|---|---|---|---|
| Yes | Yes | Yes | Yes | High |
| Yes | Yes | Yes | No | Moderate |
| Yes | Yes | No | Yes | Moderate |
| Yes | Yes | No | No | Low |
| Yes | No | Yes | Yes | Moderate |
| Yes | No | Yes | No | Low |
| Yes | No | No | Yes | Low |
| Yes | No | No | No | Extremely low |
| No | Yes | Yes | Yes | Moderate |
| No | Yes | Yes | No | Low |
| No | Yes | No | Yes | Low |
| No | Yes | No | No | Extremely low |
| No | No | Yes | Yes | Low |
| No | No | Yes | No | Extremely Low |
| No | No | No | Yes | Extremely low |
| No | No | No | No | Extremely low |

*F. Determining the Consistency factor*

The Consistency factor will be determined by taking the inputs of all five Payloads, each inputs of these five payloads is combined and Rough Set Theory is applied to obtain a corresponding consistency factor. Consistency factor determines the consistency of an Autopilot in concordance to the flight system.

Let A be 'Payload I Consistency'
Let B be 'Payload II Consistency'
Let C be 'Payload I Consistency'
Let D be 'Payload II Consistency'
Let E be 'Payload II Consistency'
Let C.F. be 'Consistency Factor'

TABLE VI

| S no. | A | B | C | D | E | C.F. |
|---|---|---|---|---|---|---|
| 1. | High | High | High | High | High | Consistent |
| 2. | Medium | High | High | High | High | Inconsistent |
| 3. | High | Medium | High | Medium | High | Consistent |
| 4. | High | High | High | Medium | High | Consistent |
| 5. | High | High | Medium | High | High | Consistent |
| 6. | Low | High | Medium | Medium | Medium | Inconsistent |
| 7. | Medium | High | High | Medium | Medium | Inconsistent |
| 8. | High | High | Medium | High | Medium | Consistent |
| 9. | High | High | Medium | Medium | High | Consistent |
| 10. | High | High | Medium | Medium | Medium | Consistent |
| 11. | Very low | High | High | High | High | Inconsistent |
| 12. | High | Low | High | High | Medium | Consistent |
| 13. | High | Medium | Low | High | Medium | Consistent |
| 14. | High | Low | High | Extremely low | Extremely low | Inconsistent |
| 15. | High | Medium | Low | High | High | Consistent |
| 16. | Low | Medium | High | Medium | Medium | Inconsistent |
| 17. | Medium | Medium | Low | Medium | High | Inconsistent |
| 18. | Low | Low | Medium | High | Medium | Inconsistent |
| 19. | High | High | Extremely low | Medium | Extremely low | Inconsistent |
| 20. | High | High | Extremely low | Medium | Medium | Inconsistent |
| 21. | High | Medium | High | Medium | Medium | Consistent |
| 22. | High | High | Extremely low | High | Extremely low | Consistent |
| 23. | High | High | Medium | High | Extremely low | Inconsistent |
| 24. | High | Medium | Medium | High | Extremely low | Consistent |
| 25. | Extremely low | Extremely low | High | Medium | Low | Inconsistent |
| 26. | High | High | Extremely low | Medium | High | Consistent |
| 27. | High | Low | Low | High | High | Inconsistent |
| 28. | High | High | Extremely low | Medium | Medium | Inconsistent |
| 29. | High | Extremely low | Medium | Medium | High | Consistent |
| 30. | Extremely low | Extremely low | Extremely low | Extremely low | Extremely low | Inconsistent |

*G. Decision Algorithm*

Payload I, Payload II, Payload III, Payload IV and Payload V are condition attributes and Consistency Factor is a Decision attribute.

*rule 1. (Payload I = high) & (Payload IV = medium) & (Payload V = high) => (Consistency Factor = Consistent);*

*rule 2. (Payload I = high) & (Payload II = high) & (Payload IV = high) => (Consistency Factor = Consistent);*

*rule 3. (Payload I = high) & (Payload II = medium) => (Consistency Factor = Consistent);*

*rule 4. (Payload I = high) & (Payload II = low) & (Payload V = medium) => (Consistency Factor = Consistent);*

*rule 5. (Payload I = high) & (Payload III = medium) & (Payload IV = medium) => (Consistency Factor = Consistent);*

rule 6. (Payload III = extremely low) & (Payload V = medium) => (Consistency Factor = Inconsistent);

rule 7. (Payload I = medium) => (Consistency Factor = Inconsistent);

rule 8. (Payload II = high) & (Payload IV = low) => (Consistency Factor = Inconsistent);

rule 9. (Payload I = extremely low) => (Consistency Factor = Inconsistent);

rule 10. (Payload I = low) => (Consistency Factor = Inconsistent);

rule 11. (Payload IV = extremely low) => (Consistency Factor = Inconsistent);

rule 12. (Payload II = low) & (Payload III = low) => (Consistency Factor = Inconsistent);

rule 13. (Payload IV = medium) & (Payload V = extremely low) => (Consistency Factor = Inconsistent);

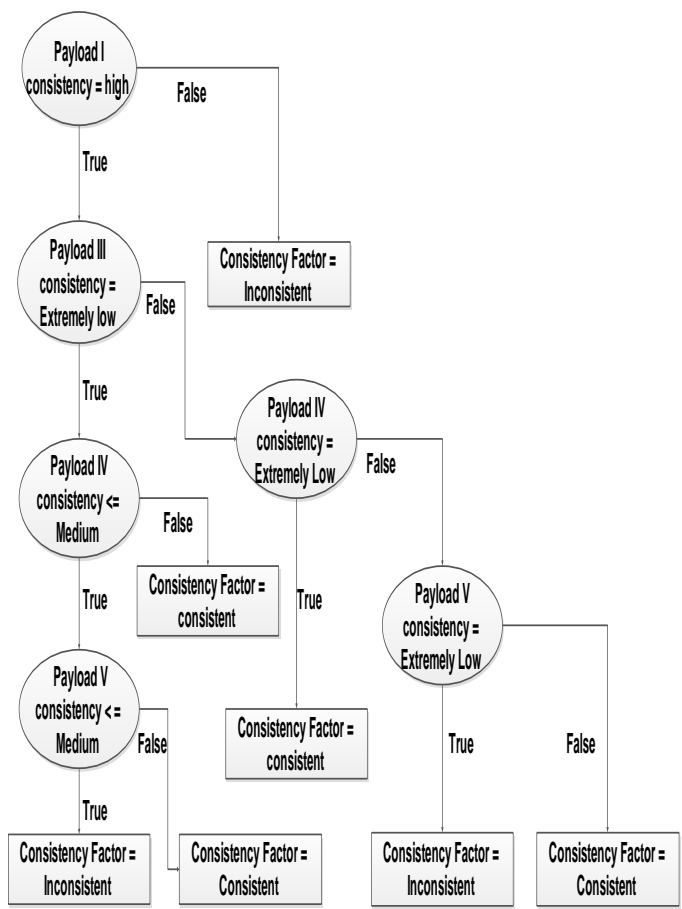

FIG. 2. DECISION TREE FOR CONSISTENCY FACTOR USING ID3 ALGORITHM.

*H. Experiments Result*

Based on the 30 sets of Training data, Rough Set Based Decision Algorithm and Decision tree using ID3 algorithm is obtained. 50 sets of Testing Data are used to validate the detection rate. Thus the accuracy of Rough Set Based Decision algorithm for determining the consistency factor can be evaluated compared to a traditional method.

TABLE VII

| Approach | Training Data set | Testing Data set | Matched Content | Detection Rate |
|---|---|---|---|---|
| Rough Set based Decision algorithm | 30 | 50 | 48 | 96% |
| ID3 based Decision Tree | 30 | 50 | 41 | 82% |

Analysis of the Experiment Data shows us that determining consistency factor using Rough Set is a more effective and accurate method. Also the simplicity of designing a code for decision rules compared to decision tree using ID3 algorithm makes Rough Set based approach a better alternative in detecting an Auto pilot system's consistency.

Analyzing the above algorithm we may infer that –

1. Payload I is the most influencing element in determining the consistency factor of an Autopilot system. Thus factors such as Roll inconsistency, Yaw inconsistency and Pitch inconsistency should have highest priority while determining Autopilot's system consistency.

2. Payload II is the least influencing element in determining the consistency factor. Inconsistency in Payload II only may allow the pilot to skip manual override during flight under usual circumstances.

V. CONCLUSION

In this paper we have implemented Rough set theory in Consistency Determination Algorithm and got more precise results. Using this approach the vast data was reduced in a more systematized and organized data. Hence, we determined a Consistency factor using Rough Sets. This Consistency Factor helps us in evaluating the current consistency of an autopilot during a flight and to alarm the pilot for manual override in case of inconsistency. Using Rough Set we also revealed, the Most and the least influencing factors while determining the consistency of an autopilot system.

REFERENCES

[1] "How airport security has changed since 9/11" Available: http://www.flightglobal.com/features/9-11/airport-security/

[2] Gary Shelly, Misty Vermaat "Discovering Computers 2011: Brief" pp. xix.

[3] "A weak excuse | Uncommon Descent" Available: http://www.uncommondescent.com/intelligent-design/a-weak-excuse/

[4] Z. Pawlak: Rough sets, International Journal of Computer and Information Sciences, 1982.

[5] L. Polkowski: Rough Sets, Mathematical Foundations, Advances in Soft Computing, Physica – Verlag, A Springer-Verlag Company, 2002.